\documentclass[10pt,twocolumn,letterpaper]{article}

\usepackage[accsupp]{axessibility}
\usepackage{cvpr}              

\usepackage{multicol,multirow}
\usepackage[dvipsnames]{xcolor}

\usepackage{algorithm}
\usepackage{algorithmic}
\usepackage{wrapfig}

\renewcommand{\eg}{\textit{e.g.}}
\renewcommand{\ie}{\textit{i.e.}}

\usepackage{adjustbox}

\definecolor{cvprblue}{rgb}{0.21,0.49,0.74}
\usepackage[pagebackref,breaklinks,colorlinks,allcolors=cvprblue]{hyperref}

\def\confName{CVPR}
\def\confYear{2025}

\title{OffsetOPT: Explicit Surface Reconstruction without Normals}

\author{Huan Lei\\
AIML, University of Adelaide\\
{\tt\small huan.lei@adelaide.edu.au}\\
{\tt\small \url{https://github.com/EnyaHermite/OffsetOPT}}
}

\begin{document}


\twocolumn[{%
\renewcommand\twocolumn[1][]{#1}%
\maketitle
\vspace{-1em}
\includegraphics[width=1.0\linewidth]{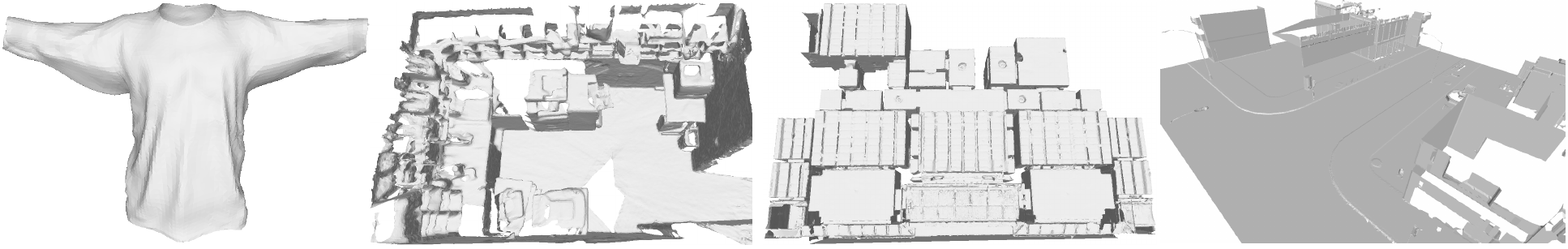}
\vspace{-1.5em}
    \captionof{figure}{We present OffsetOPT (Offset OPTimization) for explicit surface reconstruction from 3D point clouds, without the need for point normals. The prediction model is trained on synthetic meshes with supervision and then generalized to unseen point clouds through unsupervised optimization of per-point offsets. All surfaces in this figure are reconstructed using the same trained model with offset optimization. Our method achieves state-of-the-art performance in overall surface quality, sharp detail preservation, and scalability.}
\label{fig:teaser}
\vspace{1.6em}
}]

\begin{abstract}
Neural surface reconstruction has been dominated by implicit representations with marching cubes for explicit surface extraction. However, those methods typically require high-quality normals for accurate reconstruction. We propose \textbf{OffsetOPT}, a method that  reconstructs  explicit surfaces directly from 3D point clouds and eliminates the need for point normals. The approach comprises two stages: first, we train a neural network to predict surface triangles based on local point geometry, given uniformly distributed training point clouds. Next, we apply the frozen network to reconstruct surfaces from unseen point clouds by optimizing a per-point offset to maximize the accuracy of triangle predictions. Compared to state-of-the-art methods, OffsetOPT not only excels at reconstructing overall surfaces but also significantly preserves sharp surface features. We demonstrate its accuracy on popular benchmarks, including  small-scale shapes and large-scale open surfaces.
\end{abstract}
  
\section{Introduction}
\label{sec:intro}
Surface reconstruction from 3D point clouds is essential in applications across computer vision, graphics, and robotics. Traditional solutions to this problem include computational methods such as ball-pivoting~\cite{bernardini1999ball} and Delaunay triangulation~\cite{cheng2013delaunay}, along with the classic Poisson method~\cite{kazhdan2006poisson, kazhdan2013screened}, which estimates an implicit indicator function by solving a linear system. The success of Poisson surface reconstruction~\cite{kazhdan2013screened} in industry has resulted in the predominant exploration of implicit neural representations in geometric deep learning~\cite{park2019deepsdf,sitzmann2020implicit,mescheder2019occupancy,huang2023neural}. These methods generally require high-quality, oriented normals to predict a scalar field as the implicit surface representation.
Explicit surfaces are then extracted using Marching Cubes~\cite{lorensen1987marching} or its variants~\cite{ju2002dual, schaefer2004dual}.

However, Marching Cubes becomes incompatible with the unsigned distance fields (UDFs)~\cite{chibane2020neural} in open-surface reconstruction because it relies on sign changes across the surface for mesh extraction. Researchers have explored various methods~\cite{guillard2022meshudf, stella2024neural, huang2023neural} to incorporate signs into UDFs such that Marching Cubes can still be applied. In contrast to the widespread interest in implicit neural representations, neural computational methods for explicit surface reconstruction have been largely overlooked~\cite{liu2020meshing, rakotosaona2021learning, sharp2020pointtrinet, lei2023circnet}, despite their bypass of implicit representations, no reliance on point normals, and good generalization to diverse surfaces. The key limitations restricting their applications are: (i) a strong bias toward input points distributed like those from Poisson disk sampling~\cite{yuksel2015sample}, which is impractical, (ii) insufficient or inconvenient handling of edge-manifoldness in reconstructed surfaces. 

With this work, we contribute a novel neural computational method, OffsetOPT, for explicit surface reconstruction directly from 3D point clouds. The method does not rely on normals but reconstructs surfaces of both shapes and open scenes with high accuracy. More importantly, it demonstrates remarkable performance in producing edge-manifold triangles without requiring specific handling~\cite{liu2020meshing,rakotosaona2021learning, sharp2020pointtrinet,lei2023circnet}. OffsetOPT consists of two stages. First, we train a transformer-based network~\cite{vaswani2017attention} to predict triangle faces adjacent to each point based on local geometry. All of our training point clouds~\cite{koch2019abc} approximate an uniform distribution. Second, we freeze the network to reconstruct surfaces from unseen point clouds by optimizing per-point offsets to enhance the quality of triangle predictions.  

Compared to state-of-the-art methods, the proposed OffsetOPT excels not only in overall surface reconstruction but also in preserving sharp surface details. We demonstrate its accuracy on popular benchmarks, using shapes from ABC~\cite{koch2019abc}, FAUST~\cite{bogo2014faust}, and MGN~\cite{bhatnagar2019multi}, as well as large-scale scene surfaces from ScanNet~\cite{dai2017scannet}, Matterport3D~\cite{Matterport3D}, and CARLA~\cite{dosovitskiy2017carla}. Below are our main contributions: 
\begin{itemize} 
\vspace{1mm}
\item A novel neural computational method is proposed for surface reconstruction from general 3D point clouds. \vspace{1mm}
\item It does not rely on point normals or Poisson-disk sampled points for accurate surface reconstruction. \vspace{1mm}
\item Unlike the previous computational methods, it produces edge-manifold triangles without explicit handling. \vspace{1mm}
\item It outperforms existing approaches in reconstructing the overall structure and fine details of surfaces.
\end{itemize}

\section{Related Work}
\label{sec:references}
Surface reconstruction from 3D point clouds is a central research topic in geometry processing. Existing solutions for this problem generally fall into two categories: (a) computational methods that directly reconstruct explicit surfaces from point clouds; (b) implicit methods that solve for different scalar fields to represent surfaces implicitly (\eg, binary occupancy, signed distance, or unsigned distance fields). The latter must resort to Marching Cubes~\cite{lorensen1987marching} for explicit surface extraction from the scalar fields. They also depend on consistently oriented normals for reliable performance.

Traditional methods in the computational category include Alpha shapes~\cite{edelsbrunner1994three}, the ball pivoting algorithm~\cite{bernardini1999ball}, and Delaunay triangulation~\cite{cazals2004delaunay,cheng2013delaunay}, while Poisson surface reconstruction~\cite{kazhdan2006poisson,kazhdan2013screened} is a representative approach in the implicit category. Below, we briefly review neural methods advancing the two directions in geometric deep learning.

\subsection{Implicit Neural Representations}
Implicit neural representations initially focused on watertight surface reconstruction, where neural networks are used to predict occupancy fields or signed distance fields (SDFs) for the surface. Researchers have developed multiple loss functions to learn the underlying implicit surface from oriented, dense point clouds~\cite{park2019deepsdf, mescheder2019occupancy, sitzmann2020implicit, atzmon2020sal, gropp2020implicit, NeuralPull}. 

For improved reconstruction, various encoder architectures have been explored to extract more effective local features for each point, allowing the decoder to predict the occupancy status or signed distance associated with each point more accurately~\cite{chen2019learning,peng2020convolutional, wang2022dual,boulch2022poco}. Meanwhile, inspired by the success of basis functions in SPSR~\cite{kazhdan2013screened}, a number of works have applied (learnable) basis functions or neural kernels to surface reconstruction~\cite{peng2021shape, huang2022neural, williams2021neural, williams2022neural, huang2023neural}. 

Given the significance of open surface reconstruction in real-world applications, research on unsigned distance fields (UDFs) has gained increasing attention~\cite{chibane2020neural,lu2024unsigned,zhang2023surface}. However, a key limitation of UDFs is their incompatibility with Marching Cubes due to the absence of sign information. The primary solution to this issue is to convert UDFs into SDFs by introducing signs~\cite{guillard2022meshudf, stella2024neural, huang2023neural, wang2022hsdf}. Among implicit methods, NKSR~\cite{huang2023neural} is a notable contribution that holds strong inductive bias. It leverages well-oriented normals to establish SDFs for surfaces with varying topologies, including both watertight and open surfaces. 
While the method remains effective without normals, it prefers high-quality normals for superior performance. For their representativeness, we compare our reconstruction accuracy with the implicit methods, SPSR~\cite{kazhdan2013screened} and NKSR~\cite{huang2023neural}.

A known drawback of implicit methods is that they tend to oversmooth the sharp surface details~\cite{rakotosaona2021differentiable}, which we show  empirically in our experiments. It is worth noting that we focus on reconstructing surfaces from general 3D point clouds, assuming they adequately capture the underlying surface. Surface reconstruction from noisy and sparse point clouds~\cite{ouasfi2024unsupervised, chen2024neuraltps} is beyond the scope of this work.

\begin{figure*}
    \centering
    \includegraphics[width=0.99\linewidth]{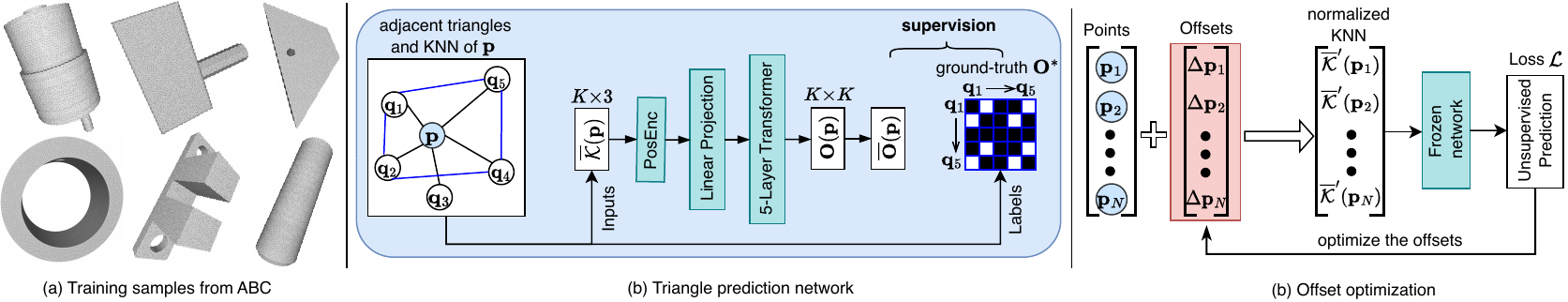}
    \vspace{-2mm}
    \caption{Overview of the proposed OffsetOPT method. (a) provides examples of training samples from the ABC dataset, showing meshes with uniformly distributed points and equilateral triangles (zoom-in for a better view). (b) is the training of our triangle prediction network in a supervised manner, where ground-truth labels are established from adjacent triangles of each point in the training meshes. The network predicts surface triangles based on KNN neighborhoods of points. (c) is the offset optimization for surface reconstruction. For a point cloud $\{{\mathbf p}_n\}$, we optimize its offsets $\{\Delta{\mathbf p}_n\}$ by backpropagating the unsupervised prediction loss through the frozen network. For each offset update during optimization, the KNN geometry used by the network is recomputed with points $\{{\mathbf p}_n + \Delta{\mathbf p}_n\}$.}
    \label{fig:offsetOPT_method}
     \vspace{-3mm}
\end{figure*} 
\subsection{Neural Computational Reconstruction}\label{related:existing_PCT}
In contrast to the popularity of implicit neural representations, neural computational reconstruction has received less attention.  Existing methods generally establish candidate triangles based on the $K$NN neighborhood information~\cite{preparata2012computational} of each point. PointTriNet~\cite{sharp2020pointtrinet} proposes candidate triangle faces using a proposal network and classifies surface triangles with a separate network. It handles edge-manifoldness between triangles explicitly in the proposal network. IER~\cite{liu2020meshing} predicts whether a candidate triangle face belongs to the reconstructed surface based on the intrinsic-extrinsic distance ratio. The complexity for establishing candidate triangles is $\mathcal{O}(K^2)$ per point, and the method is highly restricted to Poisson-disk sampled points.
DSE~\cite{rakotosaona2021learning} parameterizes the 3D neighborhood of each point onto a 2D chart, enabling the use of Delaunay triangulation~\cite{mark2008computational}. Both IER and DSE rely on a voting-based mechanism during post-processing to handle edge-manifoldness in the reconstructed surface. Different from the combinatorial formulation of previous methods, CircNet~\cite{lei2023circnet} exploits the duality between a triangle and its circumcenter to reformulate the reconstruction as a detection of triangle circumcenters, resulting in reduced complexity. Therefore, it is much faster at producing the primitive surfaces. Yet, the edge-manifoldness handling at post-processing takes time. 

The neural computational methods exhibit strong generalization to unseen point clouds across diverse shapes and scenes. In contrast, implicit neural representations are highly data-driven and constrained by their training priors when applied to unseen data. For instance, NKSR~\cite{huang2023neural} required extensive training on a combined dataset of varied shapes and scenes to establish a robust inductive bias.
Currently, a key limitation hindering the applications of computational methods is their over-reliance on input points being ideally distributed, simulating a Poisson disk sampling~\cite{yuksel2015sample}.

To justify the promise of neural computational reconstruction, we present OffsetOPT. Similar to previous approaches, it reconstructs the surface based on the local point geometry derived from their $K$NN neighborhoods. The network is trained to predict surface triangles from the $\mathcal{O}(K^2)$ candidates. Our proposed \textit{offset optimization} effectively addresses edge-manifoldness without requiring explicit handling. Besides, it extends the applicability of neural computational reconstruction to general point clouds, rather than being restricted to those produced by the Poisson method.

\vspace{-1mm}
\section{Explicit Surface Reconstruction}\label{sec:method}
\vspace{-2mm}
\noindent{\bf Overview.} Computational reconstruction preserves the input points as mesh vertices, avoiding the need for interpolating new points as in implicit methods~\cite{kazhdan2006poisson}. It directly reconstructs the explicit surface from its point cloud representation by predicting adjacent triangle faces for each point. Our method involves training a transformer-based network on point clouds with an approximately ideal uniform distribution, followed by the optimization of 3D coordinate offsets for each input point. We train the network~\cite{vaswani2017attention} with Binary Cross-Entropy (BCE) to predict triangle faces based on local point geometry, as detailed in~\S\ref{subsec:network}. The trained network is then frozen, and we optimize per-point offsets to improve the prediction accuracy, extending explicit surface reconstruction to general point clouds while achieving satisfactory edge-manifoldness (see \S\ref{subsec:offsetOPT}). Here, edge-manifoldness refers to each edge being adjacent to at most two triangles in the reconstructed surface.

\subsection{Triangle Prediction Network}
\label{subsec:network}
{\bf Local point geometry.} The local geometry of a point has been widely exploited in geometric deep learning to learn expressive feature representations~\cite{qi2017pointnet++,rakotosaona2021learning,lei2023circnet}. We therefore utilize input features derived from the  neighborhood of each point to predict surface triangles with a neural network. Let $\mathcal{K}({\mathbf p})=\{{\mathbf q}_1,{\mathbf q}_2,\dots,{\mathbf q}_{K}\}$ be the $K$NN neighborhood of point ${\mathbf p}$, where points are sorted such that $\|{\mathbf q}_k-{\mathbf p}\|\leq\|{\mathbf q}_{k+1}-{\mathbf p}\|$.
To ensure the prediction robustness across different data resolutions, we normalize the $K$NN neighborhood by the smallest non-zero distance as 
\begin{equation}\label{eq:normalized_KNN}
\overline{\mathcal{K}}({\mathbf p})=\Big\{\overline{\mathbf q}_{k}|\overline{\mathbf q}_k=\eta_0\frac{{\mathbf q}_k-{\mathbf p}}{\|{\mathbf q}_1-{\mathbf p}\|}\Big\}_{k=1}^K,
\end{equation}
with $\eta_0$ fixed as 0.01. The neural network then takes these normalized $K$NN coordinates with positional encoding~\cite{mildenhall2021nerf} to predict the surface triangles, similar to~\cite{lei2023circnet}.

\noindent\textbf{Candidate triangles.} Given the $K$ neighbors of a point~${\mathbf p}$, all combinations yield ${K \choose 2}$ candidate triangles. However, for better control of the edge-manifoldness, we predict $K^2$ triangles, represented by a symmetric matrix of size $K{\times}K$. Each entry $(i,j)$ of the matrix for point ${\mathbf p}$ corresponds to a triangle formed by $({\mathbf p},{\mathbf q}_i,{\mathbf q}_j)$, which is identical to the one in entry $(j,i)$. Triangles in the same row, such as those involving ${\mathbf q}_3$, share the edge $({\mathbf p},{\mathbf q}_3)$. The diagonal entries correspond to degenerate triangles, where two vertices are identical and hence ignored. For each edge $({\mathbf p},{\mathbf q}_i)$, we extract either two triangles or none. Our network predicts the probabilities for each candidate triangle in this matrix. For symmetric control, we predict a raw probability matrix ${\mathbf O}$ of size $K{\times}K$ and symmetrize it as $\overline{\mathbf O} = {\mathbf O} + {\mathbf O}^\intercal$. The ground-truth labels ${\mathbf O}^*$ for the supervised training of $\overline{\mathbf O}$ with BCE loss are established based on the adjacent triangles to each point in the training meshes.

\noindent\textbf{The prediction network.} Our neural network for triangle face prediction consists of a linear projection layer followed by a 5-layer transformer~\cite{vaswani2017attention}. Let $N$ be the number of points in the point cloud. The inputs to our network have dimensions $N \times K \times C_{in}$, and the outputs have dimensions $N \times K \times K$. We train the network using point clouds that approximate an uniform distribution, leveraging triangle meshes from the ABC dataset. Figure~\ref{fig:offsetOPT_method}(a) shows some  examples from our training set, which consists of simple synthetic shapes. Figure~\ref{fig:offsetOPT_method}(b) illustrates how the network is trained using the adjacent triangles and the $K$NN neighborhood of each point in the mesh.

\noindent\textbf{Triangle Extraction.} While applying the trained network to surface reconstruction, we extract triangle faces based on the predicted probabilities $\overline{\mathbf O}$ at each point. Specifically, we sort the columns of each row in $\overline{\mathbf O}$ to select the top two most likely triangle faces. Different confidence thresholds are applied to filter the selected triangles, with $p_1$ for the first triangle and $p_2$ for the second. Furthermore, we check the angle between the two triangles to avoid unwanted face folding, enforcing an angle greater than $A$. This strategy is applied to extract adjacent triangles for each point. Duplicate triangle predictions from different points are removed.

\subsection{Point Offset Optimization}\label{subsec:offsetOPT}
The trained network in \S\ref{subsec:network} performs well only on surface reconstruction from point clouds with distributions similar to the training data. To extend its applicability to general point clouds, we propose optimizing an offset for each input point, which constitutes our key contribution. The intuition behind this design is to adjust the point locations to match the preferred distribution of the network.

\noindent\textbf{Offset initialization.} We initialize the offsets of each point based on their nearest neighbors. For example, the offset of point ${\mathbf p}$ is initialized as 
\begin{equation}\label{eq:offset_init}
\Delta \mathbf{p}^0 = 0.25{\times}({\mathbf p} - {\mathbf q}_1),
\end{equation}
where ${\mathbf q}_1$ is the nearest neighbor of ${\mathbf p}$. This initialization pulls each point \textit{further apart} from its nearest neighbor. With these offsets, we re-establish the normalized $K$NN neighborhood $\overline{\mathcal{K}}(\mathbf{p})$ using the new points with added offsets (\eg, $\mathbf{p} + \Delta \mathbf{p}$). Note that we do not search those $K$NN neighborhoods again but use the original indexing and ordering. Our trained network employs the re-computed $\overline{\mathcal{K}}'(\mathbf{p})$ to calculate new input features.

\noindent\textbf{Offset optimization.} We freeze the network and compute the gradients of the offsets by minimizing the average BCE loss for triangle predictions:
\begin{equation}\label{eq:loss}
\mathcal{L}= \frac{1}{N{\times}K{\times}K}\sum_{n}\sum_{i}\sum_{j} \text{BCE}(O_{nij}).   
\end{equation}
As there are no ground-truth labels in this process, we construct pseudo-labels based on the triangle extraction strategy described in \S\ref{subsec:network}. Specifically, for each predicted $\overline{\mathbf O}$, we encourage the top two most likely triangles along each row to have labels of $1$, with the condition that the highest confidence $p_1>0.5$.

Let the raw gradient for  $\Delta{\mathbf p}$ at the $t$-th iteration be $\nabla_t(\Delta{\mathbf p})$, and $\gamma_t$ be the learning rate. The original distance between ${\mathbf p}$ and its nearest neighbor ${\mathbf q}_1$ is $d_0({\mathbf p})=\|{\mathbf p}-{\mathbf q}_1\|$. 
Based on the directions provided by the raw gradient, we control the offset update to prevent each point from drifting arbitrarily away from the surface, using the distance to its nearest neighbor. Our uncontrolled updates for each offset are computed as
\begin{equation}\label{eq:offset_uncontrolled}
\Delta{\mathbf p}^{t+1} = \Delta{\mathbf p}^{t} - \gamma_t \cdot \widetilde{\nabla}_t (\Delta{\mathbf p}), 
\end{equation}
in which 
\begin{equation}
\widetilde{\nabla}_t (\Delta{\mathbf p}) =  d_0({\mathbf p}) \cdot \frac{\nabla_t(\Delta{\mathbf p})}{\|\nabla_t(\Delta{\mathbf p})\|}.     
\end{equation}

For every uncontrolled update of the offset, we evaluate how it affects the new distance between each point and its nearest neighbor, calculated as  
\begin{equation}\label{eq:new_nn_dist}
d_{t+1} = \min_k {\|({\mathbf p}+\Delta{\mathbf p}^{t+1})-({\mathbf q}_k+\Delta{{\mathbf q}_k^{t+1}})\|},
\end{equation}
To prevent collisions and promote the repulsion between nearest-neighbor points during the offset optimization, we update the offset $\Delta{\mathbf p}$ only when $d_{t+1}{>}\frac{d_0}{2}$. Finally, the controlled updates of each offset become
\begin{equation}\label{eq:offset_controlled}
\Delta{\mathbf p}^{t+1} = \Delta{\mathbf p}^{t} - m \cdot \gamma_t \cdot \widetilde{\nabla}_t (\Delta{\mathbf p}),  
\end{equation}
with 
\begin{equation}
m = \mathbb{I}\big[d_{t+1}({\mathbf p})>0.5{\times}{d_0({\mathbf p})}\big]. 
\end{equation}

In the experiments, we apply the offset optimization with a decaying learning rate $\gamma_t$ for a specified number of iterations $T$ to reconstruct the surface effectively.

\begin{figure*}
    \centering
    \vspace{-1mm}
    \includegraphics[width=0.98\linewidth]{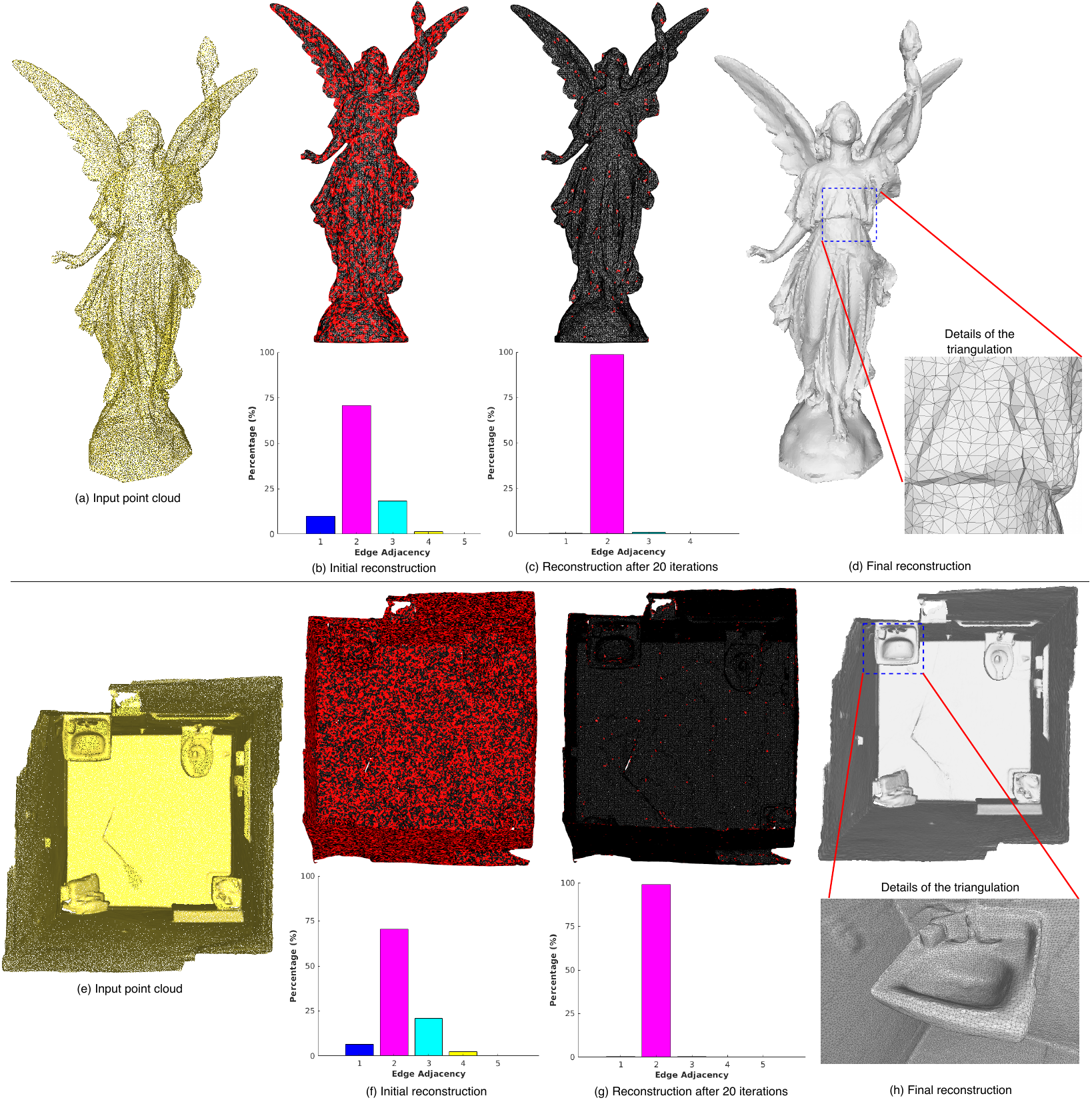}
    \vspace{-1.5mm}
    \caption{Evolution of manifold edges during OffsetOPT reconstruction. In the top row, we illustrate the reconstruction process for a shape from Thingi10k, with manifold edges in \textit{black} and non-manifold edges in \textit{red}. Given the input point cloud in (a), the initial reconstruction contains a high percentage of non-manifold edges, as shown in the histogram in (b), where `Edge Adjacency' refers to the number of faces adjacent to each edge. We note that manifold edges have an adjacency of no more than 2. After 20 iterations of offset optimization, the percentage of manifold edges increases significantly from 75\% to 99\% in (c), while the number of non-manifold edges diminishes. The final reconstructed mesh, with detailed triangulation around the belly, is shown in (d).
The bottom row shows a similar effect in scene reconstruction from ScanNet. Comparing (f) and (g), iterative offset optimization substantially increases the number of manifold edges. In (h), we display the final reconstruction with detailed triangulation around a sink.}
    \label{fig:manifold_evolve}
    \vspace{-3mm}
\end{figure*}

\noindent\textbf{Comparison to NKSR~\cite{huang2023neural}}. Our approach reconstructs surfaces by optimizing an offset for each point, making it applicable to arbitrary surfaces, even though the network is trained solely on the ABC dataset. In contrast, NKSR requires training on a diverse set of shapes and surfaces for effective generalization and relies on high-quality, well-oriented normals for optimal performance. Although normals are not used, OffsetOPT outperforms NKSR in both reconstruction accuracy and the preservation of fine details. 
\subsection{Reconstruction Application}
Although our method handles general point cloud inputs, its primary target is surface reconstruction from dense point clouds. In that case, we begin by voxelizing\footnote{voxel-based point cloud subsampling.} the dense point cloud to create a more regular input, then apply the trained network in \S\ref{subsec:network}  with offset optimization in \S\ref{subsec:offsetOPT} to reconstruct the surface. 
A recommended setting for the voxel size $v$ is to approximate the largest nearest-neighbor distance between point pairs, which is the coarsest resolution of the point cloud. We briefly summarize the proposed OffsetOPT method for surface reconstruction in Algorithm~\ref{alg:offset_opt_alg}.

For a visual understanding of how OffsetOPT promotes manifold edges during surface reconstruction, we present two examples  to demonstrate the process: one for a shape and another for a scene, as illustrated in Fig.~\ref{fig:manifold_evolve}. In both cases, the number of manifold edges increases significantly through offset optimization, resulting in high-quality triangulation details in the final reconstructed surface. 

Despite achieving notable performance in yielding edge-manifold triangles, our method may still reconstruct the surface with a small percentage of non-manifold edges. These can be removed through post-processing using functions from~\cite{lei2023circnet} or 3D libraries like Open3D~\cite{zhou2018open3d}. Given the low percentage of non-manifold edges, such post-processing is efficient and has minimal impact on the reconstruction quality. We report reconstruction accuracy of our method on the non-post-processed surfaces in the experiments.

\begin{table*}[htb]
    \centering    
    \caption{Method comparison on the \textbf{ABC} test set. `+$\hat{\mathbf{n}}$, +$\mathbf{n}$' indicate the usage of estimated and ground-truth normals, respectively; `0.1M' represents reconstruction from 0.1 million points randomly sampled from the ground-truth mesh.}\label{table:ABC}
    \vspace{-3mm}
    \small
    \begin{tabular}{l|c|c|c|c|c|c|c}
    \hline
    \multirow{3}{*}{Method}& \multicolumn{7}{c}{Surface Quality}  \\
    \cline{2-8}
    & \multicolumn{5}{c|}{overall} & \multicolumn{2}{c}{sharp}  \\
    \cline{2-8}
    & CD1($\times10^2$)$\downarrow$  & CD2($\times10^5$)$\downarrow$ & F1$\uparrow$ & NC$\uparrow$ & NR$\downarrow$ & ECD1($\times10^2$)$\downarrow$ & EF1$\uparrow$
       \\
    \hline
     ball-pivot (+$\hat{\mathbf{n}}$)&0.297 &0.684 &0.939 &0.981 &2.244 &0.782 &0.873  \\
    SPSR (+$\mathbf{n}$)&0.400 &6.081  &0.901  &0.972 &6.020 &26.160 &0.108 \\
     \cline{1-8}
     DSE&0.285&0.548 &0.949 &0.985 & 1.793 &0.538 &0.929  \\
     PointTriNet&0.288 &0.790 &0.948 &0.984 &1.931 &0.688 &0.926   \\  
     CircNet & 0.284 & 0.544 & 0.950 & 0.985 & 1.758 &0.708 &0.924    \\     
     NKSR~(+$\mathbf{n}$) & 0.370 &3.968 & 0.918 &0.978 & 5.225 &27.499 &0.097  \\
     NKSR~(+$\mathbf{n}$,~0.1M) & 0.306 &1.167 &0.938 &\textbf{0.989} &2.929 &4.152 &0.514   \\
     \cline{1-8}
     OffsetOPT (\textbf{Prop.}) & \textbf{0.283} & \textbf{0.540} & \textbf{0.951} & 0.988 & \textbf{1.318} & \textbf{0.402} & \textbf{0.941} \\
     \hline
    \end{tabular}
    \vspace{-1mm}
    \end{table*}

    \begin{table*}[htbp]  
    \centering
    \begin{minipage}{1.0\textwidth}
        \centering
        \caption{Method comparison on the \textbf{FAUST} dataset. Each point cloud contains 6,890 points by default.}
    \label{table:FAUST}
    \vspace{-3mm}
    \small
    \begin{tabular}{l|c|c|c|c|c|c|c}
    \hline
    \multirow{3}{*}{Method}& \multicolumn{7}{c}{Surface Quality} \\
    \cline{2-8}
    & \multicolumn{5}{c|}{overall} & \multicolumn{2}{c}{sharp}   \\
    \cline{2-8}
    & CD1($\times10^2$)$\downarrow$  & CD2($\times10^5$)$\downarrow$ & F1$\uparrow$ & NC$\uparrow$ & NR$\downarrow$ & ECD1($\times10^2$)$\downarrow$ & EF1$\uparrow$
       \\
    \hline
     ball-pivot (+$\hat{\mathbf{n}}$)&0.323&1.002 &0.923 &0.970 &6.037 &2.887 &0.184 \\
    SPSR (+$\mathbf{n}$)&0.427 &4.108  &0.915  &0.969 &10.269 &1.069 &0.810 \\
     \cline{1-8}
     DSE&0.218&0.307 &0.995 &0.984 &\textbf{3.910} &0.883 &0.801  
     \\
     PointTriNet&0.219&0.308 &0.995 &0.983 &4.393 &1.233 &0.807   
     \\
     CircNet  &0.221 &0.316 &0.993 &0.980 &4.557 &0.939 &0.820  \\
     NKSR (+$\mathbf{n}$) & 0.302 &0.654 &0.972 &0.973 &9.410 &2.737 &0.501  \\
     NKSR (+$\mathbf{n}$,~0.1M) & 0.227 & 0.319 &\textbf{0.997} &\textbf{0.987} &6.303 &0.970 &0.813   \\
     \cline{1-8}
     OffsetOPT (\textbf{Prop.})& \textbf{0.217} &\textbf{0.301} &0.996 &0.985 &4.038 &\textbf{0.561} & \textbf{0.896}   \\  
     \hline
    \end{tabular}
    \end{minipage}
    \vspace{2mm}  

    \begin{minipage}{1.0\textwidth}
        \centering
        \caption{Method comparison on the \textbf{MGN} open surfaces.}
     \vspace{-3mm}
    \label{table:MGN}
    \small
    \begin{tabular}{l|c|c|c|c|c|c|c}
    \hline
    \multirow{3}{*}{Method}& \multicolumn{7}{c}{Surface Quality}  \\
    \cline{2-8}
    & \multicolumn{5}{c|}{overall} & \multicolumn{2}{c}{sharp}  \\
    \cline{2-8}
    & CD1($\times10^2$)$\downarrow$  & CD2($\times10^5$)$\downarrow$ & F1$\uparrow$ & NC$\uparrow$ & NR$\downarrow$ & ECD1($\times10^2$)$\downarrow$ & EF1$\uparrow$
       \\
    \hline
    ball-pivot (+$\hat{\mathbf{n}}$)&0.462&4.917&0.844&0.974&5.803&11.847&0.083 \\
    SPSR (+$\mathbf{n}$)&1.077 &10.481  &0.402  &0.948 &12.224 &7.912 &0.137   \\
     \cline{1-8}    
     DSE&0.270&0.530&0.968&0.983&3.970&4.508&0.440 
    \\
     PointTriNet&0.272&0.562& 0.967&0.981& 4.398&5.936 &0.399   \\
     CircNet &\textbf{0.269} &0.512 &\textbf{0.968}&0.981&4.230 &\textbf{3.231} &0.486 \\
     NKSR (+$\mathbf{n}$) & 0.946 &23.263 &0.611 &0.947 &11.818 &13.815 &0.065  \\
     NKSR (+$\mathbf{n}$,~0.1M) & 0.381 &0.884 &0.891 &0.990 &4.997 &6.488 &0.441  \\     
     \cline{1-8}
     OffsetOPT (\textbf{Prop.}) & 0.278& \textbf{0.511} & 0.964 &\textbf{0.991} &\textbf{2.967} &5.447 &\textbf{0.538}  \\
     \hline
    \end{tabular}
    \end{minipage}
    \vspace{-5mm}
\end{table*}

\begin{table*}[htbp]
    \centering
        \caption{Method comparison on \textit{large-scale} open surface datasets: \textbf{ScanNet}, \textbf{Matterport3D}, and \textbf{CARLA}.}
        \small
    \label{table:Scenes}
    \vspace{-3mm}
    \begin{tabular}{l|l|c|c|c|c|c|c|c}
    \hline
    \multicolumn{2}{l|}{\multirow{3}{*}{~~~Method}}& \multicolumn{7}{c}{Surface Quality} \\
    \cline{3-9}
   \multicolumn{2}{l|}{} & \multicolumn{5}{c|}{overall} & \multicolumn{2}{c}{sharp} \\
    \cline{3-9}
   \multicolumn{2}{l|}{} & CD1($\times10^2$)$\downarrow$& CD2($\times10^5$)$\downarrow$ & F1$\uparrow$ & NC$\uparrow$ & NR$\downarrow$ &  ECD1($\times10^2$)$\downarrow$ & EF1$\uparrow$
     \\
    \hline
 \multirow{4}{*}{\rotatebox{90}{ScanNet}}   & SPSR (+$\mathbf{n}$)  & 5.428 &352.674 &0.194 &0.709 &35.904 &7.370 &0.131   \\
& NKSR (+$\mathbf{n}$)  & 0.157 &0.164 &0.997 &\textbf{0.963} &9.824 &0.618 &0.885   \\
&     NKSR & 0.423 &1.648 &0.793 &0.901 &17.699 &1.743 &0.586   \\ 
&    OffsetOPT (\textbf{Prop.}) & \textbf{0.147} &\textbf{0.136} &\textbf{1.0} &0.960 &\textbf{9.533} &\textbf{0.389} &\textbf{0.931} \\
    \hline
     \hline
 \multirow{4}{*}{\rotatebox{90}{MPort3D}}   & SPSR (+$\mathbf{n}$)  & 0.926 &28.893 &0.724 &0.830 &23.322 &2.202 & 0.344  \\
    & NKSR (+$\mathbf{n}$)  & 0.183&0.220 & 0.995 & 0.936 & 12.713 &0.619 &0.842  \\
    & NKSR  & 0.271 & 0.762 &0.939 &0.894 &18.076 & 0.903 &0.718  \\ 
    & OffsetOPT (\textbf{Prop.}) & \textbf{0.148} &\textbf{0.139}&\textbf{1.0} &\textbf{0.938} &\textbf{10.665} &\textbf{0.250} &\textbf{0.973} \\
     \hline
     \hline
\multirow{4}{*}{\rotatebox{90}{CARLA}} & SPSR (+$\mathbf{n}$)  & 4.407 &234.338 &0.121 &0.733 &30.835 &6.799 &0.062  \\     
& NKSR (+$\mathbf{n}$)  & 0.175 &0.299 &0.974 &0.953 &7.740 &0.679 & 0.903   \\
&     NKSR & 0.238 &2.682 &0.968 &0.936 &10.761 &1.013 &0.810  \\ 
&    OffsetOPT (\textbf{Prop.}) & \textbf{0.124} &\textbf{0.272} &\textbf{0.987} &\textbf{0.963} &\textbf{5.530} &\textbf{0.317} &\textbf{0.946}  \\
    \hline
    \end{tabular}
\end{table*}

\begin{figure*}[ht]
    \centering
    \includegraphics[width=0.98\linewidth]{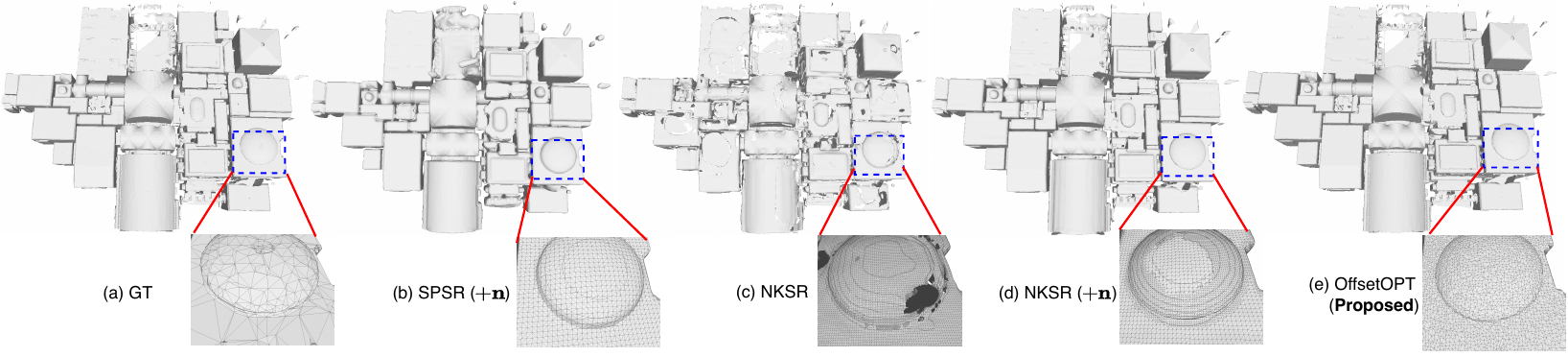}
    \vspace{-2mm}
    \caption{Comparison of different reconstruction methods on a large-scale building from Matterport3D~\cite{Matterport3D}. (a) The ground-truth surface. (b) SPSR~\cite{kazhdan2013screened} reconstruction using ground-truth normals, showing strong oversmoothing. (c) NKSR~\cite{huang2023neural} reconstruction without normals, resulting in many disconnected components. (d) The improved NKSR reconstruction with ground-truth normals. (e) Our reconstruction with OffsetOPT, which captures fine details without requiring normals. Zoomed-in triangulation details are shown for each surface.}
    \label{fig:vis_scene_compare}
    \vspace{-4mm}
\end{figure*}

\setlength{\textfloatsep}{11pt}
\begin{algorithm}[!t]
\caption{Offset optimization for surface reconstruction.}
\label{alg:offset_opt_alg} 
\begin{algorithmic}[1]
\renewcommand{\algorithmicrequire}{\textbf{Input:}}
\renewcommand{\algorithmicensure}{\textbf{Output:}}
\REQUIRE (1) A point cloud $\{{\mathbf x}_s\}_{s=1}^S$; (2) The trained network.
\ENSURE The reconstructed surface as a triangle mesh.
\STATE  Voxelize the point cloud as $\{{\mathbf p}_n\}_{n=1}^N$ with grid size $v$.
\STATE Establish the $K$NN neighborhood $\mathcal{K}({\mathbf p})$ of each point.
\STATE Initialize the corresponding offsets $\{\Delta{\mathbf p}_n^0\}_{n=1}^N$.
\FOR{iteration $t<T$}
\STATE Re-compute the normalized neighborhood  $\overline{\mathcal{K}}'(\mathbf{p})$. 
\STATE Predict triangles with the network using $\overline{\mathcal{K}}'(\mathbf{p})$. 
\STATE Compute loss $\mathcal{L}$ in Eq.~(\ref{eq:loss}) and backpropagate to get raw offset gradients $\nabla_t(\Delta{\mathbf p})$.
\STATE Compute the uncontrolled $\Delta{\mathbf p}^{t+1}$ using Eq~(\ref{eq:offset_uncontrolled}).
\STATE Check the updated NN distance $d_{t+1}$ with Eq.~(\ref{eq:new_nn_dist}).
\STATE Update the offset with the controlled Eq.~(\ref{eq:offset_controlled}).
\ENDFOR
\STATE Extract surface triangles using the strategy in \S\ref{subsec:network}.
\STATE [\textit{Optional}] Post-process for strict edge-manifoldness.
\RETURN The reconstructed triangle mesh. 
 \end{algorithmic}
 \end{algorithm}

\section{Experiments}
\textbf{Implementation details.} The transformer layers in our neural network has 64 channels and 4 attention heads. We use $K{=}50$ in the $K$NN search, and a basis level of 8 for the positional encoding. During training, we randomly sample points from each training mesh, using their adjacent triangles for ground-truth labels and KNN geometry as network inputs. The input point clouds are augmented with random rotations, scaling, and jittering. For offset optimization, the learning rate is initialized to $\gamma_0 = 0.1$ and decays by a factor of 0.7 every 10 iterations. We optimize the offsets for 100 iterations. The threshold settings for triangle extraction are $(p_1, p_2, A) = (0.8, 0.5, 120^\circ)$.

\noindent{\bf Evaluation criteria.} We assess the overall surface quality of each reconstructed mesh using symmetric Chamfer distances (CD1, CD2), F-Score (F1), normal consistency (NC), and normal reconstruction error (NR) in degrees. The quality of fine surface details is further evaluated using Edge Chamfer Distance (ECD1) and Edge F-score (EF1), following \citep{chen2022neural,lei2023circnet}. Specifically, for shape reconstruction, we sample 10\textsuperscript{5} points from both the ground-truth and reconstructed meshes, while for scene reconstruction, we use 10\textsuperscript{6} points to ensure adequate surface coverage. All meshes are re-scaled to fit within a unit sphere for metric reporting.

\noindent{\bf Training data.} The ABC dataset~\cite{koch2019abc} offers a collection of clean synthetic meshes with high-quality, nearly equilateral triangle faces. We use the 9,026 voxelized\footnote{Mesh decimation with voxel-based vertex clustering.} meshes from~\cite{lei2023circnet}, split into 25\% for training and 75\% for testing, to train and evaluate our triangle prediction network. 

\noindent{\bf Method comparison.} We compare the reconstruction quality of OffsetOPT to computational reconstruction methods, including ball-pivoting~(\cite{bernardini1999ball}), PointTriNet~\cite{sharp2020pointtrinet}, DSE~\cite{rakotosaona2021learning}, and CircNet~\cite{lei2023circnet}, as well as representative implicit neural methods, including SPSR~\cite{kazhdan2013screened} and NKSR~\cite{huang2023neural}.

\subsection{Shape Reconstruction}\label{exp:shape_reconstruct}
Trained on the ABC training set, our prediction network outperforms existing methods on the ABC test set with a simple forward pass, without offset optimization, due to the similarity in point distributions between the training and test sets. The results are reported in Table~\ref{table:ABC}.

For surface reconstruction from unseen point clouds in general, we apply the model trained on ABC with offset optimization, \ie, the proposed OffsetOPT. We validate the generalization of OffsetOPT using FAUST~\cite{bogo2014faust}, a dataset of watertight meshes of human bodies, and MGN~\cite{bhatnagar2019multi}, a dataset of open meshes of clothes. The results are reported in Table~\ref{table:FAUST}
 and Table~\ref{table:MGN}, respectively.

It can be seen that OffsetOPT consistently outperforms other approaches, especially in reconstructing sharp surface details. We note that the performance drop of NKSR on the MGN dataset is due to its tendency to close small holes in the open surfaces, such as those around the neck and legs.

For all shape reconstruction experiments, the default input points consist of the vertices of the test meshes.  We additionally sample 0.1 millon points from each test mesh to report the improved reconstruction results of NKSR. We compute estimated normals $\hat{\mathbf n}$ for ball-pivoting from the input point cloud and the ground-truth normals ${\mathbf n}$ for SPSR and NKSR from the ground-truth mesh.

\subsection{Scene Reconstruction}\label{exp:scene_reconstruct}
To demonstrate the performance of OffsetOPT in large-scale scene reconstruction, we apply it to surfaces from the test splits of ScanNet~\cite{dai2017scannet}, Matterport3D~\cite{Matterport3D}, and 10 random scenes from CARLA~\cite{dosovitskiy2017carla,huang2023neural}. In all experiments, we randomly sample one million points from the ground-truth meshes (ScanNet and Matterport3D) or dense point clouds (CARLA) as inputs. In the reconstruction with OffsetOPT, we voxelize the input points with a grid size of $2$ cm for ScanNet and $10$ cm for Matterport3D and CARLA. We compare OffsetOPT to SPSR and the state-of-the-art NKSR, with quantitative results provided in Table~\ref{table:Scenes}
 and visualized comparisons in Fig.~\ref{fig:vis_scene_compare}. It is observed that SPSR recovers the overall scene structure well but suffers from strong over-smoothing, limiting its reconstruction accuracy. In contrast, our method-without relying on normals-consistently outperforms NKSR, particularly in recovering the fine details, whereas ground-truth normals are provided for the latter. For space concern, we show additional visual comparisons in the supplementary material.
 
\vspace{-1mm}
\subsection{Ablation Study}\label{exp:ablation}
\vspace{-1mm}
\noindent\textbf{Impact of offset optimization.} Table~\ref{tab:offset_opt_impact} presents the reconstruction accuracy (CD1) and the percentage of manifold edges (\%) using our trained model  with (\textit{w}) and without (\textit{w/o}) the offset optimization in stage two. The results clearly indicate that optimizing offsets significantly promotes manifold edges, as shown in the histograms of Fig.~\ref{fig:manifold_evolve}. 
\vspace{-3.5mm}

\begin{figure}[ht]
\centering  \includegraphics[width=0.98\linewidth]{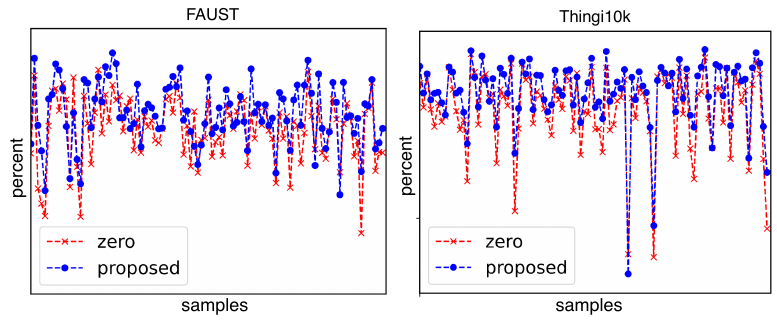}
\vspace{-2mm}
\caption{The percentage of manifold edges in the reconstructed surfaces using zero and proposed initializations, with \textit{red} for zero and \textit{blue} for proposed. Results are shown for the FAUST and Thingi10k datasets. The $x$-axis in our plots indicates the number of samples, while the $y$-axis represents the percentage.}
\label{fig:manifold_percent_init_ablation}
\vspace{-2mm}
\end{figure}

\noindent\textbf{Different offset initialization.} In Eq.~(\ref{eq:offset_init}), we slightly move each point away from its nearest neighbor for offset initialization. A more straightforward approach is to initialize all offsets as zeros. Figure~\ref{fig:manifold_percent_init_ablation} illustrates the percentage of manifold edges in reconstructed surfaces using different initializations for the FAUST (\S\ref{exp:shape_reconstruct}) and Thingi10k~\cite{zhou2016thingi10k} datasets. This experiment uses 100 shapes from Thingi10k, with point clouds consisting of 0.1 million points randomly sampled from the ground-truth meshes. See Table~\ref{table:offset_init_ablation} in the supplementary for the reconstruction accuracies using the two initializations. We adopt the proposed initialization as it yields slightly better performance.

\begin{table}[!t]
    \centering
    \scriptsize
    \caption{Impact of offset optimization.}\label{tab:offset_opt_impact}
    \vspace{-3.2mm}
    \setlength{\tabcolsep}{2.8pt}
    \begin{tabular}{l|c|c|c|c|c}
    \hline
   offset & ABC & FAUST & MGN & ScanNet & MP3D  \\
     \hline
   \textit{w/o} & \textbf{0.283} (\textbf{99\%}) & 0.221 (82\%)& 0.280 (88\%)& 0.154 (74\%)& 0.164 (78\%) \\
     \hline
   \textit{w} & - & \textbf{0.217} (\textbf{98\%})& \textbf{0.278} (\textbf{99\%})& \textbf{0.147} (\textbf{99\%})& \textbf{0.148} (\textbf{99\%}) \\
     \hline
    \end{tabular}
     \vspace{-2mm}
\end{table}

\vspace{-2mm}
\section{Limitation}
\vspace{-1mm}
Although the proposed OffsetOPT demonstrates remarkable improvements in producing manifold edges compared to prior computational reconstruction methods, a small percentage of non-manifold edges may still occur. Future work will aim to address this limitation more comprehensively. Additionally, our prediction network utilizes a transformer architecture, which, while achieving superior reconstruction accuracy, is less efficient than the graph convolutional networks~\cite{lei2023circnet}. Exploring the combination of graph convolutions with transformers in future work could better balance accuracy and efficiency in surface reconstruction, while potentially enhancing method robustness. Currently, our approach takes an average of 15.67 seconds and 13.83 seconds to reconstruct each shape in FAUST and MGN, respectively, and approximately 6, 12, and 9 minutes to reconstruct each scene in ScanNet, Matterport3D, and CARLA, using a single NVIDIA GeForce RTX 4090 GPU.

\section{Conclusion}
We have introduced a novel neural computational method, OffsetOPT, for reconstructing explicit surfaces directly from 3D point clouds. Unlike the implicit methods, it does not require point normals. OffsetOPT operates in two stages: it first trains a triangle prediction network based on local point geometry to reconstruct surfaces from ideally distributed points, and then applies the trained model to optimize per-point offsets for accurate surface reconstruction from general point clouds. Our approach outperforms existing methods in both overall reconstruction accuracy and the preservation of sharp surface features. We validate its effectiveness across a diverse range of benchmark datasets, including both small-scale shapes and large-scale indoor and outdoor surfaces, demonstrating its ability to achieve high-quality surface reconstructions without relying on normals.

\section{Acknowledgments}
This research was supported by the Centre for Augmented Reasoning (CAR) at the Australian Institute for Machine Learning (AIML), University of Adelaide.

{
    \small
    \bibliographystyle{ieeenat_fullname}
    \bibliography{main}
}

\clearpage
\setcounter{page}{1}
\maketitlesupplementary
\section{Additional ablation study}
\label{sec:offset_init}
\vspace{-2mm}
Table~\ref{table:offset_init_ablation} presents the comparison of reconstruction accuracy for different offset initializations discussed in \S\ref{exp:ablation}.

\begin{figure}[ht]
    \centering
    \hspace{-5mm}
    \includegraphics[width=0.48\textwidth]{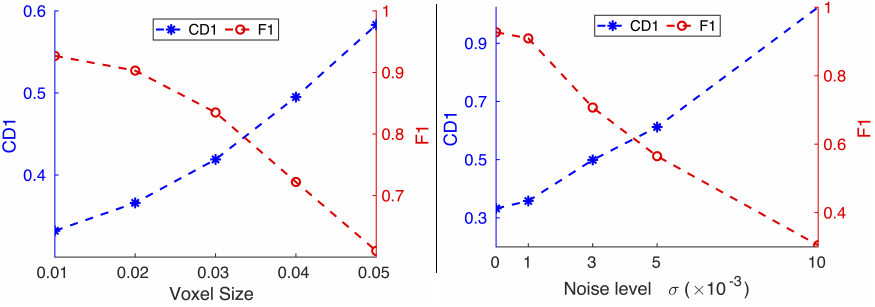}
    \vspace{-2mm}
    \caption{Ablation studies. (left) Effects of voxel sizes on reconstruction quality. (right) Method robustness to data noise.}
     \label{fig:ablation}
     \vspace{-2mm}
\end{figure}

\noindent\textbf{Voxel sizes.} We examine how voxel size affects the reconstruction accuracy using 100 shapes from Thingi10k~\cite{zhou2016thingi10k}. Each input point cloud consists of 0.1 million randomly sampled points. We normalize the points  to fit within a unit sphere, and reconstruct each sample with OffsetOPT using voxel sizes ranging from 0.01 to 0.05. We plot in Fig.~\ref{fig:ablation}(left) the variations in CD1 and F1 score to show how the reconstruction accuracy drops as voxel size increases.

\noindent\textbf{Data noise.} Using the same Thingi10k point clouds, we further evaluate the effect of data noise on OffsetOPT. Specifically, we add varying levels of Gaussian noise (\textit{i.e.}, $\sigma{\in}\{0,0.001, 0.002, 0.003, 0.005, 0.1\}$) to create noisy point clouds. Figure~\ref{fig:ablation}(right) illustrates our performance as noise increases in the shapes. We note that computational reconstruction methods use noisy points directly as mesh vertices in the reconstructed surface, resulting in reduced robustness to noise compared to implicit methods. 

\noindent\textbf{KNN influence.} We always sort the KNN points by distances (\S\ref{subsec:network}). In the training set, since distant points receive the label $0$ in the ground-truth matrices $\mathbf{O}^*$,  the model learns to discourage problematic edges or triangles itself in stage one. To illustrate, we train another model using $K{=}100$ and test it on FAUST and MGN. The results are similar to those of $K{=}50$, shown in Table~\ref{tab:knn_choice}. We suggest $50$ for efficiency. 
\begin{table}[!h]
    \centering
    \vspace{-3mm}
    \caption{Different $K$ in KNN for reconstruction accuracy.}\label{tab:knn_choice}
    \vspace{-3mm}
    \scriptsize
    \setlength{\tabcolsep}{2.pt}
    \begin{tabular}{l|c|c|c|c|c|c|c|c|c|c}
    \hline
    \multirow{2}{*}{KNN}  & \multicolumn{5}{c|}{FAUST} & \multicolumn{5}{c}{MGN} \\
    \cline{2-11}
    & CD1$\downarrow$ & CD2$\downarrow$ &F1$\uparrow$  & NC$\uparrow$ & NR$\downarrow$ & CD1$\downarrow$ & CD2$\downarrow$ &F1$\uparrow$  & NC$\uparrow$ & NR$\downarrow$  \\
     \hline
     $K{=}50$ & 0.217 & 0.301 & 0.996& 0.985 & 4.038 & 0.278 &\textbf{0.511} &0.964 & 0.991&2.967 \\
     \hline
     $K{=}100$ & 0.217 & 0.301 & 0.996& 0.985 & \textbf{3.802} & 0.278 &0.512 &0.964 & 0.991&\textbf{2.803} \\
     \hline
    \end{tabular}
    \vspace{-5mm}
\end{table}
\vspace{-2mm}
\section{Large-scale Reconstruction}
\vspace{-2mm}
\textbf{Chunk processing.} The OffsetOPT approach updates per-point offsets using gradients computed from the entire input point cloud. In large-scale reconstructions, where memory limitations prevent processing all data at once, we divide the input into manageable chunks. Gradients from each chunk are accumulated, and offsets are updated after such accumulation. This chunking strategy is flexible, allowing random splits based on point number and adapting to GPU memory capacity, without the need for careful sub-volume division. Algorithm~\ref{alg:offsetOPT_chunk_alg} shows our pseudocode for such  processing in PyTorch.
It is worth noting that, for convenience, we normalize every input point cloud into a unit sphere in all reconstructions, regardless of the original data scale.

\rmfamily
\begin{algorithm}[!t]
\caption{Offset Gradient Accumulation in Pytorch.}
\label{alg:offsetOPT_chunk_alg} 
\begin{algorithmic}[1]
\renewcommand{\algorithmicrequire}{\textbf{Input:}}
\renewcommand{\algorithmicensure}{\textbf{Output:}}
\REQUIRE Number of points per-chunk $I$.
\ENSURE Raw gradients of offsets.
\STATE Compute the number of chunks {\color{CornflowerBlue}\textit{$C=\lceil \frac{N}{I}\rceil$}}.
\STATE  {\color{CornflowerBlue}\textit{optimizer.zero\_grad()}}.
\FOR{iteration $c<C$}
\STATE  Get chunk points indexed from {\color{CornflowerBlue}$I{\times}c$} to {\color{CornflowerBlue}$I{\times}(c{+}1)$}.
\STATE Forward pass their $K$NN inputs to get {\color{CornflowerBlue}\textit{chunk\_logits}}. 
\STATE {\color{CornflowerBlue}\textit{chunk\_loss = BCE(chunk\_logits)}}.
\STATE Accumulate gradients with {\color{CornflowerBlue}\textit{chunk\_loss.backward()}}.
\ENDFOR 
\RETURN {\color{CornflowerBlue}offsets.grad}.
 \end{algorithmic}
 \end{algorithm}

\begin{wrapfigure}{r}{0.11
\textwidth}
\vspace{-3mm}
\hspace{-3mm}\includegraphics[width=0.12\textwidth]{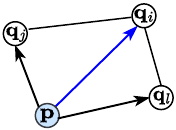}
\end{wrapfigure}
\noindent\textbf{Angle between adjacent triangles.}  
We compute the angle between two triangles sharing an edge in \S\ref{subsec:network} as follows. Let the two triangles be 
$(\mathbf{p},\mathbf{q}_i,\mathbf{q}_j)$ and $(\mathbf{p},\mathbf{q}_i,\mathbf{q}_l)$, $i\neq j\neq l$. We compute the angle $A$  between them as 
\begin{align}
  &{\mathbf N}_j   = (\mathbf{q}_i-\mathbf{p}){\bf \times} (\mathbf{q}_j-\mathbf{p}),\\
  & {\mathbf N}_l  = (\mathbf{q}_i-\mathbf{p}){\bf \times} (\mathbf{q}_l-\mathbf{p}),\\
 & A = \arccos\left( \left\langle \frac{{\mathbf N}_j}{\|{\mathbf N}_j\|}, \frac{{\mathbf N}_l}{\|{\mathbf N}_l\|} \right\rangle \right).
\end{align}
For planar surfaces, the angle between any two adjacent triangles sharing an edge will be $180^\circ$.

\vspace{-2mm}
\section{More Visualizations}
\begin{figure*}[b]
    \centering  
    \includegraphics[width=1.0
    \textwidth]{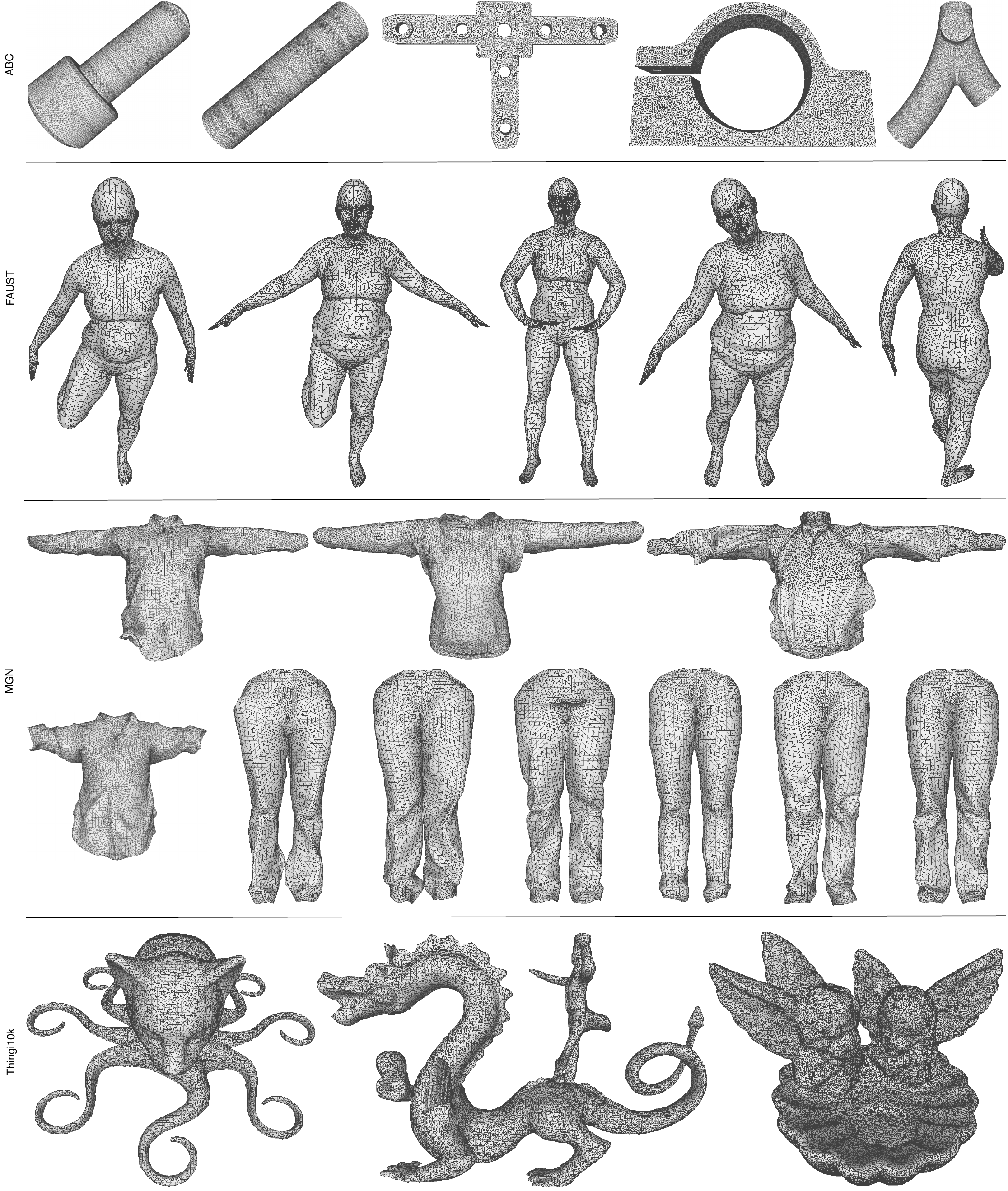}
    \caption{Examples of reconstructed shape surfaces (with wireframes) from ABC, FAUST, MGN, and Thingi10k using the proposed OffsetOPT. For the ABC dataset, results are obtained with a single forward pass of the trained network, without offset optimization, as mentioned in the main paper. OffsetOPT demonstrates surface reconstruction with satisfactory triangulations for all these shape datasets.}
    \label{fig:shape_triangles_more}
    \vspace{-8mm}
\end{figure*}

\begin{figure*}[b]
    \centering  
    \includegraphics[width=1.0
    \textwidth]{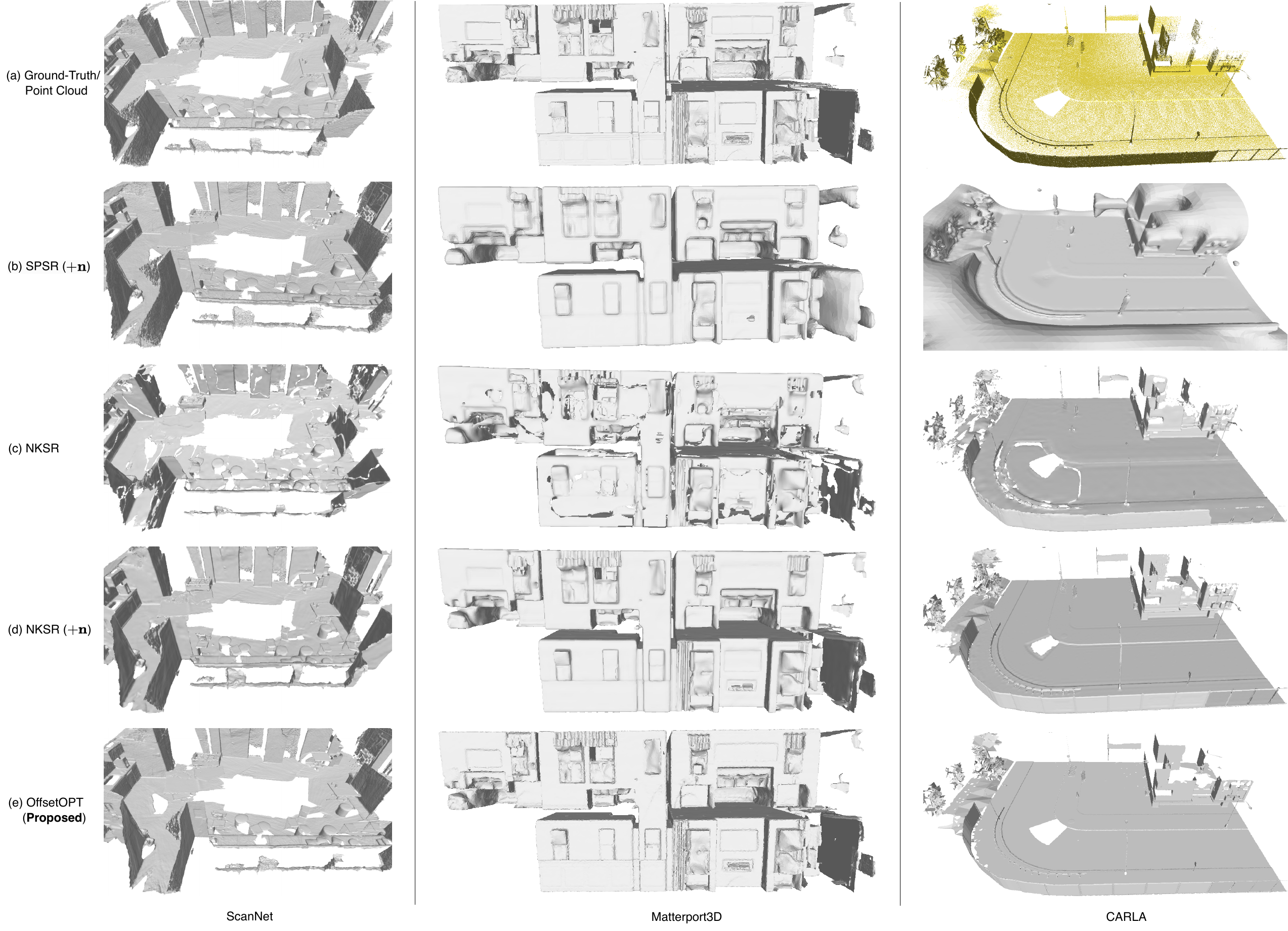}
    \caption{More comparison of different methods on reconstructing the large-scale scenes from ScanNet, Matterport3D, and CARLA.}
    \label{fig:scene_compare_more}
    \vspace{-8mm}
\end{figure*}

\vspace{-2mm}
Figure~\ref{fig:scene_triangles_more} compares the reconstructed scene surfaces of SPSR, NKSR, and the proposed OffsetOPT, with triangulation details. Consistently, we observe that SPSR produces surfaces with undesired over-smoothing, while NKSR requires ground-truth normals to perform comparably to our method.
Figure~\ref{fig:shape_triangles_more} shows our reconstructed shape surfaces with wireframes for ABC, FAUST, MGN, and Thingi10k, demonstrating the satisfactory triangulation capability of OffsetOPT. Figure~\ref{fig:scene_compare_more} provides further comparison of these methods on large-scale scene reconstruction.

\begin{table*}[hb]
    \centering
    \begin{minipage}{\textwidth}
    \centering
    \caption{Reconstruction accuracy of different offset initializations.}
    \label{table:offset_init_ablation}
    \vspace{-2mm}
    \begin{tabular}{l|l|c|c|c|c|c|c|c}
    \hline
    \multirow{3}{*}{Dataset} & \multirow{3}{*}{Initialization}& \multicolumn{7}{c}{Surface Quality}  \\
    \cline{3-9}
   & & \multicolumn{5}{c|}{overall} & \multicolumn{2}{c}{sharp} \\
    \cline{3-9}
   & & CD1($\times10^2$)$\downarrow$& CD2($\times10^5$)$\downarrow$ & F1$\uparrow$ & NC$\uparrow$ & NR$\downarrow$ &  ECD1($\times10^2$)$\downarrow$ & EF1$\uparrow$
     \\
      \hline
    \multirow{2}{*}{FAUST}& Proposed & 0.217 &0.301 &0.996 &0.985 &4.038 &\textbf{0.561} & \textbf{0.896}
    \\ 
    &zero & 0.217 &0.301 &0.996 &0.985 &\textbf{3.835} &0.584 &0.889
    \\
    \hline 
    \multirow{2}{*}{Thingi10k}& Proposed & 0.332 &\textbf{0.699} &\textbf{0.927} &\textbf{0.984} &\textbf{5.523} &\textbf{4.252} &\textbf{0.478}
    \\ 
    &zero & 0.332 & 0.701 &0.926 & 0.983 &5.534 &4.484 &0.476 
    \\
    \hline 
    \end{tabular}
    \vspace{6mm}
    \end{minipage}
    \begin{minipage}{\textwidth}
    \centering
    \includegraphics[width=1.0\textwidth]{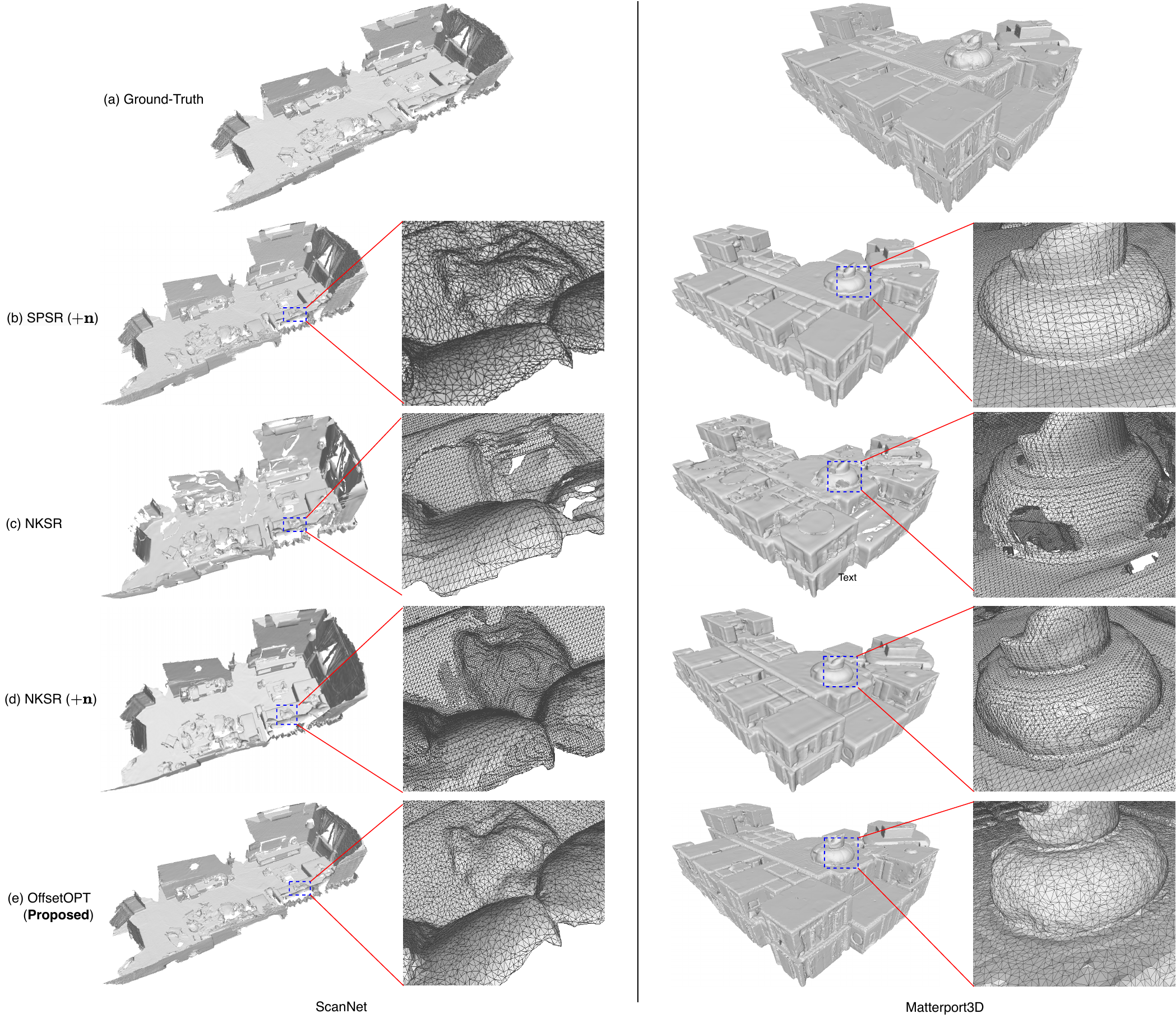}
    \caption{Reconstructed surfaces of SPSR, NKSR, and OffsetOPT, with triangulation details for a scene from ScanNet and a scene from Matterport3D. Our method recovers surfaces with sharp features, while NKSR requires ground-truth normals to achieve comparable quality.}
    \label{fig:scene_triangles_more}
    \end{minipage}
\end{table*}

\end{document}